\documentclass[a4paper, 10pt, conference]{ieeeconf}      

\IEEEoverridecommandlockouts                              
                                                          
\usepackage[para,online,flushleft]{threeparttable}
\usepackage[linesnumbered,ruled,vlined]{algorithm2e}
\usepackage{booktabs}
\usepackage{tabularray}
\usepackage{multirow}
\usepackage{adjustbox}
\usepackage{graphicx}
\usepackage{siunitx}
\usepackage[utf8]{inputenc}
\usepackage[T1]{fontenc}
\usepackage{amsfonts}
\usepackage{gensymb}
\usepackage[acronyms, shortcuts]{glossaries}

\usepackage{amsthm}

\usepackage{epsfig} 
\usepackage{amsmath} 
\usepackage{amssymb}  
\usepackage{siunitx}
\usepackage{subcaption}
\usepackage{caption}
\usepackage{color}
\usepackage[table]{xcolor}
\usepackage{soul}
\usepackage{theoremref}
\usepackage[noadjust]{cite}
\usepackage{amsthm}
\usepackage{mathtools}
\DeclarePairedDelimiter{\ceil}{\lceil}{\rceil}

\usepackage{hyperref}
\hypersetup{
    colorlinks=true,
    linkcolor=blue,
    filecolor=magenta,      
    urlcolor=blue,
    pdftitle={Overleaf Example},
    pdfpagemode=FullScreen,
    }

\usepackage[noabbrev,capitalise]{cleveref}

\newtheorem*{theorem*}{Theorem}

\newacronym[longplural={mixed-integer convex programs}]{micp}{MICP}{mixed-integer convex program}
\newacronym[longplural={mixed-integer nonlinear programs}]{minlp}{MINLP}{mixed-integer nonlinear program}
\newacronym[longplural={mixed-integer optimal control problems}]{miocp}{MIOCP}{mixed-integer optimal control problem}

\newacronym[longplural={second-order cone programs}]{socp}{SOCP}{second-order cone program}
\newacronym[longplural={mixed-integer second-order cone programs}]{misocp}{MISOCP}{mixed-integer second-order cone program}

\newacronym[longplural={quadratically constrained convex programs}]{qcqp}{QCQP}{quadratically constrained convex program}
\newacronym[longplural={mixed-integer quadratically constrained convex programs}]{miqcqp}{MIQCQP}{mixed-integer quadratically constrained quadratic program}

\newacronym[longplural={quadratic programs}]{qp}{QP}{quadratic program}
\newacronym[longplural={sequential quadratic programs}]{sqp}{SQP}{sequential quadratic program}
\newacronym[longplural={mixed-integer quadratic programs}]{miqp}{MIQP}{mixed-integer quadratic program}

\newacronym[longplural={Markov decision processes}]{mdp}{MDP}{Markov decision process}
\newacronym{mpc}{MPC}{model predictive control}
\newacronym{pwa}{PWA}{piecewise affine}

\newacronym[longplural={recurrent neural networks}]{rnn}{RNN}{recurrent neural network}
\newacronym[longplural={long short-term memory}]{lstm}{LSTM}{long short-term memory}
\newacronym[longplural={gated recurrent units}]{gru}{GRU}{gated recurrent unit}

\newacronym{b&b}{B\&B}{branch-and-bound}

\newcommand{\coco}{CoCo}

\definecolor{wheat}{rgb}{0.96,0.87,0.70}

\title{\LARGE\bf Evaluating Data-driven Performances of Mixed Integer Bilinear Formulations for Book Placement Planning}
\author{Xuan Lin$^{1}$, Gabriel I.~Fernandez$^{1}$, and Dennis W.~Hong$^{1}$
\thanks{$^{1}$X. Lin, Gabriel I.~Fernandez, and Dennis W.~Hong are with the Robotics and Mechanisms Laboratory, Department of Mechanical and Aerospace Engineering, University of California, Los Angeles, CA 90095, USA.
        {\tt\small \{maynight,gabriel808,dennishong\}@ucla.edu}}
}

\begin{document}
\maketitle
\thispagestyle{empty}
\pagestyle{empty}

\begin{abstract}
Mixed integer bilinear programs (MIBLPs) offer tools to resolve robotics motion planning problems with orthogonal rotation matrices or static moment balance, but require long solving times. Recent work utilizing data-driven methods has shown potential to overcome this issue allowing for applications on larger scale problems. To solve mixed-integer bilinear programs online with data-driven methods, several re-formulations exist including mathematical programming with complementary constraints (MPCC), and mixed-integer programming (MIP). In this work, we compare the data-driven performances of various MIBLP reformulations using a book placement problem that has discrete configuration switches and bilinear constraints. The success rate, cost, and solving time are compared along with non-data-driven methods. Our results demonstrate the advantage of using data-driven methods to accelerate the solving speed of MIBLPs, and provide references for users to choose the suitable re-formulation.
\end{abstract}
%
%

\section{Introduction}
\label{Sec:introduction}
Optimization-based methods are useful tools for solving robotic motion planning problems. Typical approaches such as mixed-integer convex programs (MICPs) \cite{deits2014footstep, lin2019optimization}, nonlinear or nonconvex programs (NLPs) \cite{DaiValenzuelaEtAl2014,winkler2018gait,shirai2020risk} and~\acp{minlp} \cite{soler2011route} offer powerful tools to formulate these problems. However, each has its drawbacks. NLPs tend to converge to local optimal solutions. In practice, local optimal solutions can have bad properties, such as inconsistent behavior as they depend on initial guesses. To introduce discrete variables, NLPs require complementary constraints \cite{PosaCantuEtAl2014}, which is numerically challenging as it violates the majority of Constraint Qualifications established for standard nonlinear optimization \cite{luo1996mathematical}. Mixed-integer programs (MIPs) explicitly treat the discrete variables, which is usually solved with Branch and Bound \cite{boyd2007branch}, or Benders Decomposition \cite{geoffrion1972generalized, lin2023generalized}. MIP solvers seek global optimal solutions with more consistent behavior than NLP solvers, but can require impractically long solving times for problems with a large number of integer variables \cite{lin2021designing}. MINLPs, such as mixed-integer bilinear programs (MIBLPs), incorporate both integer variables and nonlinear (bilinear) constraints, hence, very expressive. Unfortunately, we lack efficient algorithms to tackle MINLPs. As a result, it is difficult to implement most of the optimization schemes mentioned above online.

\begin{figure}[t!]
\centering
\includegraphics[width=0.35\textwidth]{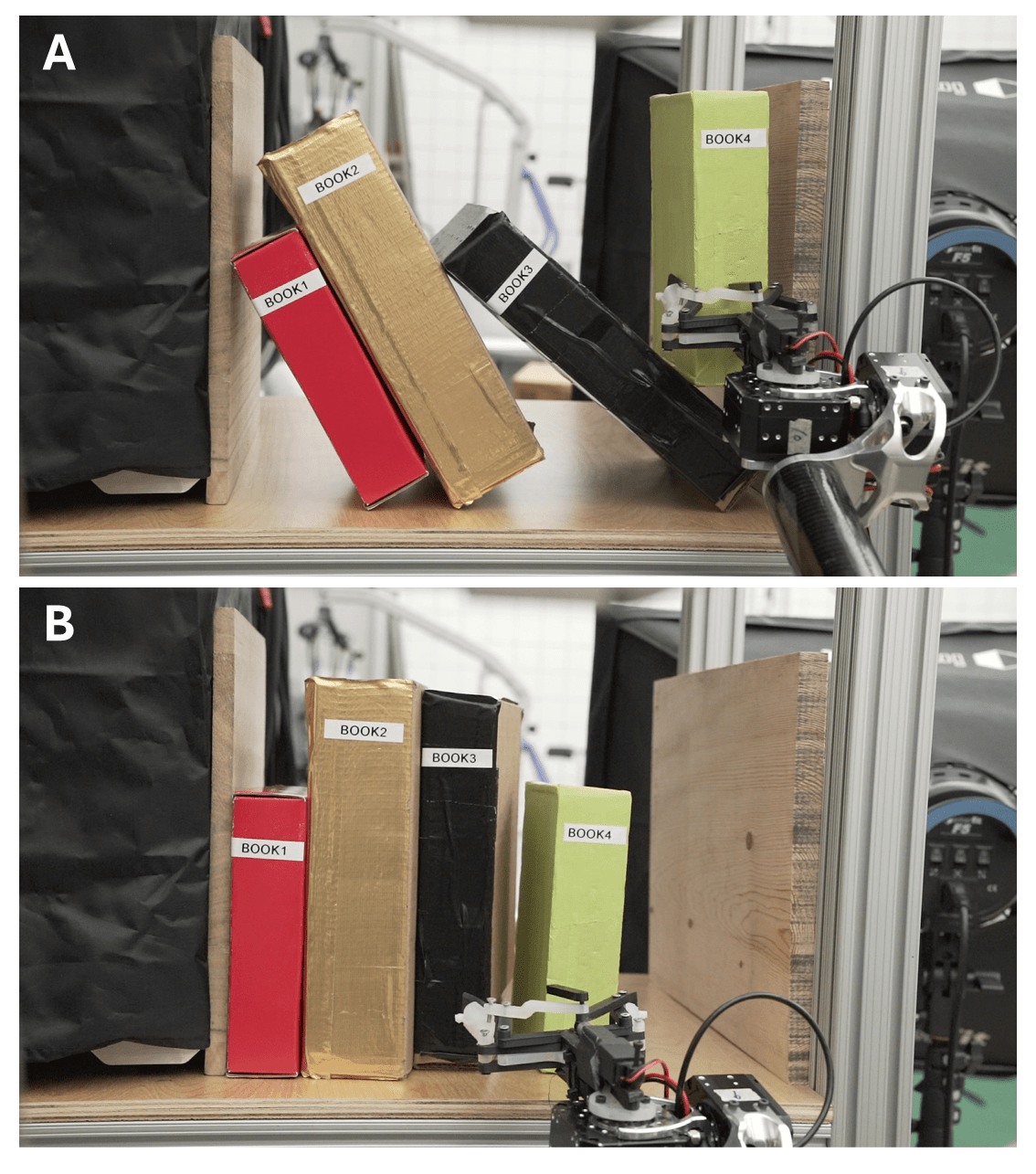}
\caption{Two snapshots of a book placement trajectory.}
\label{Fig:Stow}
\end{figure}

Recently, researchers have started to investigate machine learning methods to gather problem-specific heuristics and speed up the discrete or nonlinear optimization solvers. Standard algorithms to solve MIPs such as branch-and-bound rely on heuristics to quickly remove infeasible regions. Learning methods can be used to acquire better heuristics. For example, \cite{nair2020solving} used graph neural networks to learn heuristics. \cite{tang2020reinforcement} used reinforcement learning to discover efficient cutting planes. On the other hand, data can be collected to learn and solve specific problems \cite{ZhuMartius2020,CauligiCulbertsonEtAl2021}. In \cite{CauligiCulbertsonEtAl2021}, the authors proposed~\coco{} which turns prediction of binary variables into classification problems, where each set of binary variables has a unique class label. For online solving, the neural network proposes candidate binary solutions, reducing the problems into convex programs. Similar strategy is used by \cite{hauser2016learning} which used K-nearest-neighbor to predict initial guesses for NLPs, followed by a searching scheme to satisfy the nonlinear constraints. 


For multi-agent motion planning problems, in addition to bilinear constraints such as orthogonal rotation matrices, agents can operate under various modes or configurations encoded with discrete variables. The discrete configuration switches with bilinear constraints lead to challenging MIBLPs. There exist several approaches to re-formulate the MIBLP into either an MICP or an NLP, such that the aforementioned data-driven methods can directly apply to speed up the solver, including:

\begin{enumerate}
    \item Turning binary variables into continuous variables with complementary constraints, hence the problem becomes an MPCC \cite{park2018semidefinite}, and data-driven methods such as \cite{hauser2016learning, lin2022reduce} can directly apply.
    \item Turning bilinear constraints into mixed-integer linear constraints using McCormick envelopes \cite{gupte2013solving}, hence the problem becomes MICP, and data-driven methods such as \cite{ZhuMartius2020, CauligiCulbertsonEtAl2021} can directly apply.
\end{enumerate}

In this paper, we ask the following question:

\textit{Given a certain amount of problems solved offline as warm-start data, which re-formulation can solve the original MIBLP faster and more reliably? How much data is needed to achieve such performances?}

To get a comprehensive answer of this question, we introduce the book placement problem for a benchmark. Given a bookshelf with several books on top, an additional book needs to be placed on the shelf with minimal disturbance to the existing books. Fig. \ref{Fig:Stow} shows two snapshots of the trajectory to insert one book into the shelf while interacting with existing books. The bookshelf problem works well as a good benchmark because it: 1) is an MINLP that can be converted to an MICP problem with hundreds of integer variables, 2) can easily be scaled to push algorithms to their limit, and 3) has practical significance where data can reasonably be gathered, such as in the logistics industry.
 
To summarize, our contributions are as follows:
\begin{enumerate}
    \item Formulate the book placement problem as an MIBLP and solve it with data-driven methods. 
    \item Benchmark the data-driven performance of different MIBLP re-formulations on the bookshelf problem.
\end{enumerate}

\section{Related Works}
\label{Sec:related_work}

\subsection{Parametric Programming}
Parametric programming is a technique to build a function that maps to the optimization solutions from varying parameters of the optimization problem \cite{fiacco1983introduction, pistikopoulos2002line}. Previous works have investigated parametric programming on linear programming \cite{gal2010postoptimal}, quadratic programming \cite{bemporad2002explicit}, mixed-integer nonlinear programming \cite{dua1999algorithms}. As an implementation of parametric programming on controller design, explicit MPC \cite{bemporad2002explicit, tondel2003algorithm} tried to solve the model-predictive control problem offline and stored the active sets. When computed online, the problem parameter can be used to retrieve the active sets. \cite{bemporad2002explicit} solved a constrained linear quadratic regulator problem with explicit MPC and proved that the active sets are polyhedrons. Therefore, the parameter space can be partitioned using an algorithm proposed by \cite{dua2000algorithm}. However, when the problem is non-convex, the active sets do not in general form polyhedrons. For non-convex problems, previous works \cite{hauser2016learning, tang2019data, ZhuMartius2020} directly stored the data points and picked out warm-start using non-parametric learning such as K-nearest neighbor (KNN), and solved the online formulation. On the other hand, modern learning techniques such as neural networks can learn an embedding of a larger set of parameters that maps to the solutions \cite{CauligiCulbertsonEtAl2021}. One advantage of using neural network methods is the capability to generalize to out-of-distribution situations that are not included in the training set \cite{power2022variational}. 

\subsection{Mixed-integer Programming}
On the other hand, mathematical programs with complementary constraints (MPCC) models discrete modes through continuous variables with complementary constraints. Complementary constraints enforce a pair of variables such that if one of them is non-zero, the other one should be zero. This constraint is traditionally hard to solve and is sensitive to initial guesses. Algorithms such as time-stepping \cite{anitescu1997formulating}, pivoting \cite{drumwright2015rapidly}, central path methods \cite{kojima1991unified} are proposed to resolve complementarity. In the robotics community, complementary constraints are typically used to optimize over gaits for trajectory optimization \cite{PosaCantuEtAl2014, zhang2021transition} or control with implicit contacts \cite{cleac2021fast} where contact forces and distance to the ground are complementary with each other. On the other hand, complementary constraints can also be used to model binary variables \cite{park2018semidefinite}.

\subsection{Mathematical Programming with Complementary Constraints}
Mathematical programs with complementary constraints (MPCC) model discrete variables by turning them into continuous variables with an additional complementary constraint. Complementary constraints enforce a pair of variables such that if one of them is non-zero, the other one should be zero. This constraint is challenging to solve as it violates the majority of Constraint Qualifications established for standard nonlinear optimization \cite{luo1996mathematical}. Algorithms such as time-stepping \cite{anitescu1997formulating}, pivoting \cite{drumwright2015rapidly}, and central path methods \cite{kojima1991unified} were proposed to resolve this challenge. In the robotics community, complementary constraints have been used to optimize over gaits for trajectory optimization \cite{PosaCantuEtAl2014, zhang2021transition} or control with implicit contacts \cite{cleac2021fast} where ground contact forces and distance to the ground are complementary with each other. On the other hand, complementary constraints were also used to model binary variables \cite{park2018semidefinite}.

\subsection{Item Arrangement Planning}
Optimal object arrangement and re-arrangement problems have been studied by a large amount of previous literature. Using the task and motion planning framework, previous works \cite{krontiris2015dealing, nam2020fast, wang2021uniform} planned a sequence of motion primitives to arrange the objects into the target configurations. \cite{chen2023optimal} used MIP models to optimize the item placements on a shelf to minimize the access time. \cite{ohnishi2022mip} also used MIP models to generate book relocation plans in a library. \cite{cheong2020relocate} proposed an algorithm to determine where to
relocate objects during the rearrangement process. Those works focused on item arrangements while ignoring their interactions during the manipulation process. Along with the stow project published by Amazon Science \cite{Stow}, several recent works investigated how items interact with each other within the limited stowing space, and how to generate a motion plan considering object configurations and contact forces \cite{lin2022reduce, chen2023predicting}. However, this task remains challenging due to complex item interactions such as friction, and the variety of item configurations.

\section{Book Placement Problem Setup}
\label{Sec:problem_setup}
\begin{figure*}[!t]
		\centering
		\hspace*{-0.25cm}
		\includegraphics[scale=0.52]{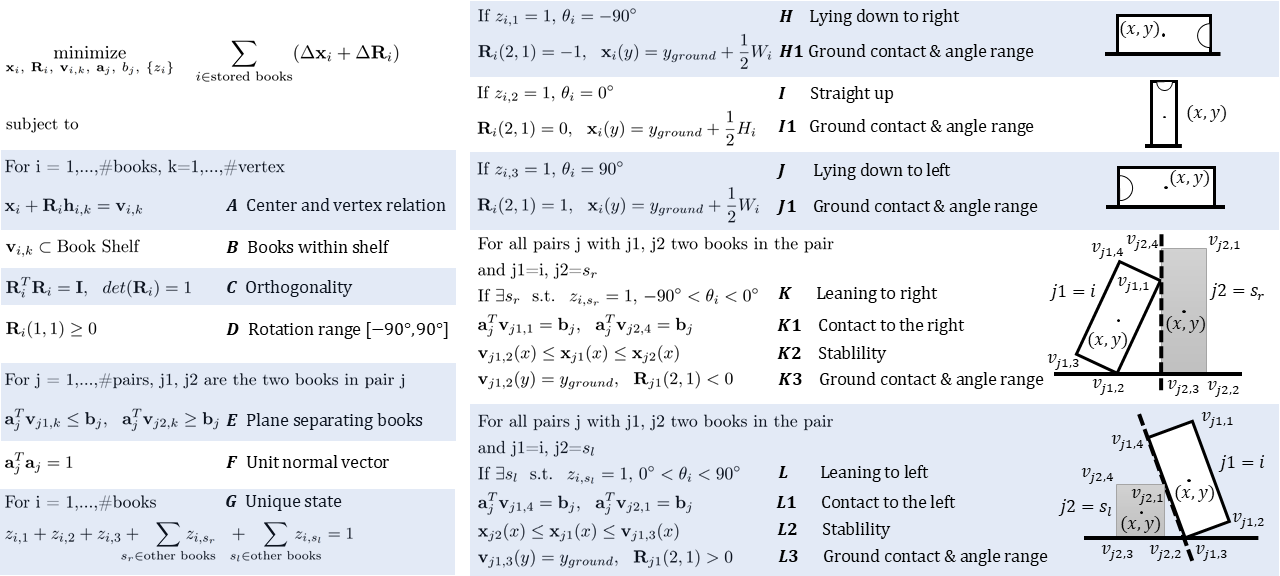}
		\caption {Complete formulation of the book placement problem.}
		\label{Fig:Complete_formulation}
\end{figure*}

Assume a 2D bookshelf with limited width $W \in \mathbb{R}$ and height $H \in \mathbb{R}$ contains rectangular books where book $i$ has width $W_{i} \in \mathbb{R}$ and height $H_{i} \in \mathbb{R}$ for $i=1,...,N-1$. A new book, $i=N$, is to be inserted into the shelf. The bookshelf contains enough books in various orientations, where in order to insert book $N$, the other $N-1$ books may need to be moved, i.e., optimize for minimal movement of $N-1$ books. This paper focuses on inserting one book using a single robot motion during which the book is always grasped. The problem settings can be extended to a sequence of motions with re-grasping.




Fig. \ref{Fig:Complete_formulation} shows the constraints, variables and objective function for the bookshelf problem. The variables that characterize book $i$ are: position $\boldsymbol{x}_{i}  \in \mathbb{R}^2 = [x_{i}, y_{i}]$ and angle $\theta_{i} \in \mathbb{R}$ about its centroid. $\theta_{i}=0$ when a book stands upright. The rotation matrix is: $\textbf{R}_{i} \in \mathbb{R}^{2 \times 2} = [cos(\theta_{i}), \ -sin(\theta_{i}); \ sin(\theta_{i}), \ cos(\theta_{i})]$. Let the 4 vertices of book $i$ be $\boldsymbol{v}_{i,k} \in \mathbb{R}^2$, $k=1,2,3,4$. The constraint \textbf{A} in Fig. \ref{Fig:Complete_formulation} shows the linear relationship between $\boldsymbol{x}_{i}$ and $\boldsymbol{v}_{i,k}$, where $\boldsymbol{h}_{i,k} \in \mathbb{R}^2$ is the constant offset vector from its centroid to vertices. Constraint \textbf{B} enforces that all vertices of all books stay within the bookshelf, a linear constraint. Constraint \textbf{C} enforces the orthogonality of the rotation matrix, a bilinear (non-convex) constraint. Constraint \textbf{D} enforces that the angle $\theta_{i}$ stays within $[-90\degree, 90\degree]$, storing books right side up.

To ensure that the final book positions and orientations do not overlap with each other, separating plane constraints are enforced. For convex shapes, the two shapes do not overlap with each other if and only if there exists a separating hyperplane $\boldsymbol{a}^{T}\boldsymbol{x}=b$ in between for $\boldsymbol{a} \in \mathbb{R}^2$ representing the plane normal direction and $b \in \mathbb{R}$ representing plane offset \cite{boyd2004convex}. That is, for any point $\boldsymbol{p}_{1} \in \mathbb{R}^2$ inside shape 1 then $\boldsymbol{a}^{T}\boldsymbol{p}_{1} \leq b$, and for any point $\boldsymbol{p}_{2} \in \mathbb{R}^2$ inside shape 2 then $\boldsymbol{a}^{T}\boldsymbol{p}_{2} \geq b$. This is represented by constraint \textbf{E}. Constraint \textbf{F} enforces $\boldsymbol{a}$ to be a normal vector. Both \textbf{E} and \textbf{F} are bilinear constraints.



Finally, we need to assign configurations to each book $i$. For each book, it can be standing straight up, laying down on its left or right, or leaning towards left or right against some other book, as shown in the far right column in Fig. \ref{Fig:Complete_formulation}. For each book $i$, we assign $n_z=3$ binary variables $\boldsymbol{z}_{i} \in \{0, 1\}^{n_z}$. If book $i$ stands upright ($z_{i,2}=1$) or lays flat on its left ($z_{i,1}=1$) or right ($z_{i,3}=1$), constraints \textbf{I1} or \textbf{J1} or \textbf{H1} are enforced, respectively. If book $i$ leans against another book (or left/right wall which can be treated as static books) on the left or right, constraints in \textbf{K} and \textbf{L} are enforced, respectively. To this end checks need to indicate the contact between books. By looking at the right column in Fig. \ref{Fig:Complete_formulation}, we can reasonably assume that the separating plane $\boldsymbol{a}^{T}\boldsymbol{x}=b$ always crosses vertex 1 of the book on the left and vertex 4 of the book on the right. This is represented by bilinear constraints, \textbf{K1} and \textbf{L1}. In addition, the books need to remain stable given gravity. Constraints \textbf{K2} and \textbf{L2} enforce that a book is stable if its $x$ position stays between the supporting point of itself (vertex 2 if leaning rightward and vertex 3 if leaning leftward) and the $x$ position of the book that it is leaning onto. Lastly, constraint \textbf{K3} and \textbf{L3} enforce that the books have contact with the \emph{ground}. For practical reasons, we assume that books cannot stack onto each other, i.e, each book has to touch the \emph{ground} of bookshelf at at least one point. We note that constraints in \textbf{H}, \textbf{I}, \textbf{J}, \textbf{K}, and \textbf{L} can be formulated as mixed-integer linear constraints using big-M formulation \cite{vielma2015mixed}. Any contact conditions between pairs of books may also be added into this problem as long as it can be formulated as mixed-integer convex constraint. Overall, this is a problem with integer variables, $\boldsymbol{z}_{i}$, and bilinear constraints \textbf{C}, \textbf{E}, \textbf{F}, \textbf{K1}, and \textbf{L1}, hence, an MIBLP problem.

Practically, this problem presents challenges for retrieving high-quality solutions. If robots were used to store books, the permissible solving time would be several seconds. Several non-data-driven approaches have been benchmarked in this paper to solve this mixed-integer bilinear problem: fix one set of bilinear variables and solve MICP \cite{posa2015stability}, nonlinear ADMM approach \cite{shirai2022simultaneous, lin2022multi}, or directly apply off-the-shelf MINLP solvers such as BONMIN \cite{bonami2007bonmin}. These non-data-driven approaches struggle to meet the requirements.

\section{Data-driven Algorithm} 
\label{Sec:learning_algorithm}
Assume that we are given a set of problems parametrized by $\boldsymbol{\Theta} \in \mathbb{R}^p$ that is drawn from a distribution $D(\boldsymbol{\Theta})$. We seek to solve a parameterized version of MIBLP \cite{fischetti2020branch}:

\begin{subequations}
\begin{align}
& \underset{\boldsymbol{x}, \ \boldsymbol{z}}{\text{minimize}} \ ||\boldsymbol{x}-\boldsymbol{x}_g(\boldsymbol{\Theta})||^2_{\boldsymbol{Q}}\\
\text{s.t.} \ \ & \boldsymbol{A}(\boldsymbol{\Theta}) \boldsymbol{x} \leq \boldsymbol{h}(\boldsymbol{\Theta}) \label{eqn:formulation-1} \\
& l_i \leq x[i] \leq u_i \ \ \ i=1,..., dim(x) \label{eqn:formulation-2} \\
& x[j] \in \{0, 1\} \ \  \ \ j \in \mathcal{J} \label{eqn:formulation-3} \\
& x[r_s] = x[p_s]x[q_s] \  s=1,...,S \label{eqn:formulation-4}
\end{align}
\label{Eqn:General_formulation}
\end{subequations}


Where $\boldsymbol{x} \in \mathbb{R}^{dim(x)}$ denotes both continuous and binary variables. $\boldsymbol{x}_g$ denotes the target values of $\boldsymbol{x}$. $\mathcal{J}$ gives an index set for binary variables. Constraints ~\eqref{eqn:formulation-1} is mixed-integer linear, meaning if the binary variables are relaxed into continuous variables $x[j] \in [0, 1] \forall j \in \mathcal{J}$, ~\eqref{eqn:formulation-1} becomes linear. Equality constraints in ~\eqref{eqn:formulation-1} are omitted as they can be turned into two inequality constraints from opposite directions.  Constraints ~\eqref{eqn:formulation-4} are bilinear.

As mentioned in section \ref{Sec:introduction}, we compare the data-driven performance of two re-formulations: converting a mixed-integer bilinear program to either an MIQP or an MPCC, and solve it with data-driven methods. If the learning agent with the problem parameter $\boldsymbol{\Theta}$ can provide a list of binary variables, the problem reduces to a convex optimization permitting convex solvers. On the other hand, if the learning agent provides a good initial guess for the variables involved in complementary constraints, the feasibility of solving the problem significantly increases. We first collect the dataset. Let the dataset of size $D$ be a set of tuples $\{(\boldsymbol{\Theta}_i, \boldsymbol{x}_i)\}$, $i=1,...,D$. We construct $\{\boldsymbol{\Theta}_i\}$ from uniform samples, then solve each $\boldsymbol{\Theta}_i$ with a non-data-driven approach to get $\boldsymbol{x}_i$. The dataset is then used to learn candidate binary variables or initial guesses for continuous variables. In the following sections, we describe the implementation details for each online formulation.

\subsection{MPCC Re-formulation}
One re-formulation of ~\eqref{Eqn:General_formulation} is to turn all the binary variables into continuous variables with complementary constraints, such that the MIBLP formulation becomes MPCC and can be solved through NLP solvers. We remove constraint ~\eqref{eqn:formulation-3}, and impose the equivalent complementary constraint from \cite{park2018semidefinite}:

\begin{equation}
    x[j](1-x[j]) \leq \epsilon \ \ j \in \mathcal{J}
\end{equation}

Where $\epsilon>0$ is a small constant to relax the constraint for better solver performance. We have also tried other implementations in \cite{stein2004continuous} and get similar or worse performances.

With the dataset, the learner samples multiple initial guesses $\boldsymbol{x}$ for the online problem $\boldsymbol{\Theta}$, using the K-nearest-neighbor approach. The MPCC is then solved using the sampled $\boldsymbol{x}$ as initial guesses, followed by a search scheme until nonlinear constraints are satisfied. This algorithm is given by Algorithm \ref{Algorithm:MPCC}.

\begin{algorithm}
\caption{MPCC Online Solving Procedure}
\label{Algorithm:MPCC}
\KwIn{$\{(\boldsymbol{\Theta}_i, \boldsymbol{x}_i)\}$, $\boldsymbol{\Theta}$}
{Find the K-nearest-neighbors $\boldsymbol{\Theta}_1$, ..., $\boldsymbol{\Theta}_K$ in $\{\boldsymbol{\Theta}_i\}$, sorted in order of increasing Euclidean distance $d(\boldsymbol{\Theta}, \boldsymbol{\Theta}_k)$. Retrieve $\boldsymbol{x}_1$, ..., $\boldsymbol{x}_K$.}\\
\For{$k=1,...,K$}{
{Solve the MPCC re-formulation of ~\eqref{Eqn:General_formulation} using applicable NLP solvers with $\boldsymbol{x}_k$ as initial guess.} \\
\If{successful}{
\Return{$\boldsymbol{x}$}\\}
}
\Return{\text{nil}}
\end{algorithm}

\subsection{MIQP Re-formulation}
Another re-formulatoin is to convert bilinear constraints into mixed-integer linear constraints. The key idea is for each constraint ~\eqref{eqn:formulation-4}, we partition the range of $x[p_s]$ and $x[q_s]$ into small intervals, then approximate the constraint within each interval using the linear McCormick envelope relaxation \cite{mccormick1976computability}. For $x[r_s] = x[p_s]x[q_s]$, we set in total $n_{p_s}$ small intervals for $x[p_s]$ as $[l_{p_s}, x^1_{p_s},, ..., x^{n_{p_s}-1}_{p_s}, u_{p_s}]$, and in total $n_{q_s}$ small intervals for $x[q_s]$ as $[l_{q_s}, x^1_{q_s},, ..., x^{n_{q_s}-1}_{q_s}, u_{q_s}]$. We then adopt the special ordered set of type 2 (sos2) formulation introduced by \cite{vielma2011modeling, dai2019global} that used additional continuous variables $\boldsymbol{\alpha}_s \in \mathbb{R}^{n_{p_s}+1}$, $\boldsymbol{\beta}_s \in \mathbb{R}^{n_{q_s}+1}$, $\boldsymbol{\gamma}_s \in \mathbb{R}^{(n_{p_s}+1) \times (n_{q_s}+1)}$, and $n_\delta$ additional binary variables $\boldsymbol{\delta} \in \{0, 1\}^{n_\delta}$, $n_\delta = \ceil{log_{2}((n_{p_s}+1) \times (n_{q_s}+1))}$, to formulate $x[r_s] = x[p_s]x[q_s]$ into mixed-integer linear constraints:

\begin{equation}
\begin{aligned}
    & x[p_s] = \sum_i{\alpha_{s}[i] x^i_{p_s}}, \ \ x[q_s] = \sum_i{\beta_{s}[i] x^i_{q_s}}\\
    & x[r_s] = \sum_i\sum_j \gamma_{s}[i,j] x^i_{p_s} x^j_{q_s} \\
    & \sum_j \gamma_{s}[i,j] = \alpha_{s}[i], \ \ \sum_i \gamma_{s}[i,j] = \beta_{s}[j], \ \ \gamma_{s}[i,j] \geq 0 \\
    & \boldsymbol{\alpha}_s, \boldsymbol{\beta}_s \ \ \text{are in sos2 (enforced by $n_{\delta}$ binary variables)}
\end{aligned}
\label{eqn:sos2_envelopes}
\end{equation}

We refer readers to \cite{vielma2011modeling} for more details of sos2 enforcement using binary variables. Despite the relatively smaller number of binary variables due to $log_{2}(\cdot)$ formulation, approximating nonlinear constraints with mixed-integer linear constraints still generates a large number of integer variables when high approximation accuracy is desired. As a result, the formulation suffers from extended solving time without good warm-starts. We used offline solved problems to train the learner. Online, the learner provides a) complete set of binary variables to fix the binary variables in ~\eqref{eqn:formulation-3}; b) complete set of continuous variables, such that one can locate the intervals for $x[p_s]$, $x[q_s]$, and replace constraints in ~\eqref{eqn:formulation-4} by linear constraints. Therefore, the problem becomes QP permitting fast solvers. For the learner, we have implemented both K-nearest neighbor approach \cite{ZhuMartius2020}, and the neural-network approach \cite{CauligiCulbertsonEtAl2021}. This KNN approach is detailed Algorithm \ref{Algorithm:MIP}. For the neural-network approach, we refer readers to \cite{CauligiCulbertsonEtAl2021} for details of network structure, training, and online sampling. After the neural network samples $\boldsymbol{x}_1, ..., \boldsymbol{x}_K$, the method to turn MIQPs into QPs is identical to Algorithm \ref{Algorithm:MIP}.

\begin{algorithm}
\caption{MIQP Online Solving Procedure}
\label{Algorithm:MIP}
\KwIn{$\{(\boldsymbol{\Theta}_i, \boldsymbol{x}_i)\}$, $\boldsymbol{\Theta}$}
{Find the K-nearest-neighbors $\boldsymbol{\Theta}_1$, ..., $\boldsymbol{\Theta}_K$ in $\{\boldsymbol{\Theta}_i\}$, sorted in order of increasing Euclidean distance $d(\boldsymbol{\Theta}, \boldsymbol{\Theta}_k)$. Retrieve $\boldsymbol{x}_1$, ..., $\boldsymbol{x}_K$.}\\
\For{$k=1,...,K$}{
\For{$s=1,...,S$}{
{Find the interval $[x^i_{p_s}, x^{i+1}_{p_s}]$ and $[x^j_{q_s}, x^{j+1}_{q_s}]$ that $x_k[p_s]$, $x_k[q_s]$ stays within, replace ~\eqref{eqn:formulation-4} with linear constraints:}\\
{\begin{align*}
    & x[p_s] = \alpha_{s}[i] x^i_{p_s} + \alpha_{s}[i+1] x^{i+1}_{p_s} \\
    & x[q_s] = \beta_{s}[j] x^j_{q_s} + \beta_{s}[j+1] x^{j+1}_{q_s} \\
    & x[r_s] = \gamma_{s}[i,j] x^i_{p_s} x^j_{q_s} + \gamma_{s}[i+1,j] x^{i+1}_{p_s} x^j_{q_s} \\
    & + \gamma_{s}[i,j+1] x^i_{p_s} x^{j+1}_{q_s} + \gamma_{s}[i+1,j+1] x^{i+1}_{p_s} x^{j+1}_{q_s} \\
    & \gamma_{s}[i,j] + \gamma_{s}[i,j+1] = \alpha_{s}[i] \\
    & \gamma_{s}[i,j] + \gamma_{s}[i+1,j] = \beta_{s}[j] \\
    & \alpha_{s}[i] + \alpha_{s}[i+1] = 1 \\
    & \beta_{s}[j] + \beta_{s}[j+1] = 1 \\
    & \alpha_{s}[i], \alpha_{s}[i+1], \beta_{s}[j], \beta_{s}[j+1] \geq 0 \\
    & \gamma_{s}[i,j], \gamma_{s}[i+1,j], \gamma_{s}[i,j+1], \gamma_{s}[i+1,j+1] \geq 0
\end{align*}}
}
{Fix $x[j]=x_k[j] \ \forall \ j \in \mathcal{J}$} \\
{Solve the resultant QP using applicable solvers.} \\
\If{successful}{
\Return{$\boldsymbol{x}$}\\}
}
\Return{\text{nil}}
\end{algorithm}

\subsection{ADMM Algorithm for MINLPs}
\label{Sec:ADMM_algorithm}
For benchmark, we implemented a non-data-driven ADMM algorithm designed for MINLPs \cite{shirai2022simultaneous}. The idea is to separate the MIBLP into a MIQP formulation and an NLP formulation with the exact copies of variables but different constraints, and iterate between them until the consensus is reached. The MIQP formulation contains constraints $\textbf{A}, \textbf{B}, \textbf{D}, \textbf{G}, \textbf{H1}, \textbf{I1}, \textbf{J1}, \textbf{K2}, \textbf{K3}, \textbf{L2}, \textbf{L3}$, while missing bilinear constraints $\textbf{C}, \textbf{E}, \textbf{F}, \textbf{K1}, \textbf{L1}$. The NLP formulation contains constraints $\textbf{A}, \textbf{B}, \textbf{C}, \textbf{D}, \textbf{E}, \textbf{F}, \textbf{H}, \textbf{I}, \textbf{J}, \textbf{K}, \textbf{L}$, while missing the mode constraint $\textbf{G}$ saying $\boldsymbol{z}$s are integers. The formulation is elaborated in Eqn. \ref{eqn:MIP_NLP}.

\begin{equation}
\begin{aligned}
& \underset{\boldsymbol{x}_{i}, \ \boldsymbol{R}_{i}, \ \boldsymbol{v}_{i,k}, \ \boldsymbol{a}_{j}, \ b_{j}, \ z_{i}}{\text{minimize}} \quad f_{obj} \\
& \text{subject to} \\
& \text{\textbf{Mixed integer linear constraints:}} \\
& \text{A, B, D, G, H1, I1, J1, K2, K3, L2, L3} \\
& \text{\textbf{Nonlinear constraints:}} \\  
& \text{A, B, C, D, E, F, H, I, J, K, L}
\end{aligned}
\label{eqn:MIP_NLP}
\end{equation}

The ADMM iterates the following 3 steps: first it solves the MIQP formulation in (\ref{eqn:MIP_NLP}), then solves the NLP formulation in (\ref{eqn:MIP_NLP}). Finally it updates the residual variables. The process continues till convergence. Let $\boldsymbol{x}_{1} \in \mathbb{R}^{n_x}$ and $\boldsymbol{x}_{2} \in \mathbb{R}^{n_x}$ be the copies of the variables in formulation (\ref{eqn:MIP_NLP}). Let $\boldsymbol{w} \in \mathbb{R}^{n_x}$ be the residual variables to achieve consensus. $\gamma$ be a constant. $\boldsymbol{G}$ are the weights. The first step is:

\begin{equation*}
\begin{aligned}
& \underset{\boldsymbol{x}_{1}}{\text{minimize}} \quad ||\boldsymbol{x}_{1}^{i}-\boldsymbol{x}_{2}^{i}+\boldsymbol{w}^{i}||_{\boldsymbol{G}_{k}} \\
& \text{s.t.  Mixed-integer constraints in ~\eqref{eqn:MIP_NLP}} 
\end{aligned}
\end{equation*}

The next step solves the NLP:
\begin{equation*}
\begin{aligned}
& \underset{\boldsymbol{x}_{2}}{\text{minimize}} \quad ||\boldsymbol{x}_{2}^{i}-(\boldsymbol{x}_{1}^{i+1}+\boldsymbol{w}^{i}))||_{\boldsymbol{G}_{k}} \\
& \text{s.t.  Nonlinear constraints in ~\eqref{eqn:MIP_NLP}} 
\end{aligned}
\end{equation*}

Finally, we update the residual variables and weights:

\begin{equation}
\begin{aligned}
    \boldsymbol{w}^{i+1} = \boldsymbol{w}^{i} + \boldsymbol{x}_{1}^{i} - \boldsymbol{x}_{2}^{i} \\
    \boldsymbol{G}_{k+1} = \gamma\boldsymbol{G}_{k} \\
    \boldsymbol{w}^{i+1} = \boldsymbol{w}^{i} / \gamma
\end{aligned}
\end{equation}

\section{Experiment} 
\subsection{Experiment Setup}
\label{Sec:experiment_setup}
We place 3 books inside the shelf where 1 additional book is to be inserted. Grids are assigned to the variables involved in the bilinear constraint \textbf{C}, \textbf{E}, \textbf{F}, \textbf{K1}, \textbf{L1}: $\boldsymbol{R}_{i}(\theta_{i})$, $\boldsymbol{a}_{j}$ and $\boldsymbol{v}_{i,k}$. These variables span a 48-dim space. $\boldsymbol{\Theta} \in \mathbb{R}^{17}$ includes the center positions, angles, heights and widths of stored books and height and width of the book to be inserted.

For MIQP re-formulation, Table \ref{Tab:variable_range} lists the number of grids for each variable involved in the bilinear constraints. The rotation angles $\theta_{i}$, which includes $\boldsymbol{R}_{i}$, are gridded at $\frac{\pi}{8}$ intervals. Elements in $\boldsymbol{a}_{j}$ are gridded on 0.25 intervals. Elements in $\boldsymbol{v}_{i,k}$ are gridded at intervals $\frac{1}{4}$ the shelf width $W$ and height $H$. Our MIQP re-formulation results in 130 integer variables in total, out of which 18 variables represent discrete configurations, and the rest comes from bilinear constraints. For MPCC re-formulations, a small relaxation of $\epsilon = 10^{-8}$ on the right hand side of the complementary constraints is used to increase the chance of getting a feasible solution.

\begin{table}[]
\centering
\caption{Segmentations of Non-convex Variables}
\begin{tabular}{l|c|c}
\hline
\multicolumn{1}{c|}{variable}   & range             & \# of segmentations \\ \hline
Item orientation $\theta$ (rad) & {[}-pi/2, pi/2{]} & 8                   \\ \hline
Separating plane normal $a$     & {[}-1, 1{]}       & 8                   \\ \hline
Vertex x position $v_{x}$ (cm)  & {[}-9, 9{]}       & 4                   \\ \hline
Vertex y position $v_{y}$ (cm)  & {[}0, 11{]}       & 4                   \\ \hline
\end{tabular}
\label{Tab:variable_range}
\end{table}

The problem instances are generated using a 2-dim simulated environment of books on a shelf. Initially, 4 randomly sized books are arbitrarily placed on the shelf, and then 1 is randomly removed and regarded as the book to be inserted. Contrary to the sequence, the initial state with 4 books represents one feasible (not necessarily optimal) solution to the problem of placing a book on a shelf with 3 existing books. This guarantees that all problem instances are feasible. Since this problem can be viewed as high-level planning for robotic systems, the simulated data is sufficient. For applications outside of the scope of this paper real-world data may be preferable in this pipeline.

All methods are tested on 400 randomly sampled bookshelf problems using the same method mentioned above. The solving process is done on a Core i7-7800X 3.5GHz × 12 machine. For NLPs, we used off-the-shelf solver KNITRO with its default interior point method. For MIQPs, we used off-the-shelf solver Gurobi. For QPs, we used off-the-shelf solver OSQP that runs ADMM. We have also tried other solvers such as IPOPT for NLPs, and Gurobi for QPs and got similar or worse results. All results are listed in table \ref{Tab:benchmark}. We omit the error bars as they are consistent among the results: success rates are within $\pm 3\%$, averaged solving times are within $\pm 10ms$, and averaged objective values are within $\pm 400$.

\begin{table*}
\centering
\caption{Benchmark Results}
\label{Tab:benchmark}
\begin{tblr}{
  column{even} = {c},
  column{3} = {c},
  column{5} = {c},
  column{7} = {c},
  column{9} = {c},
  cell{1}{5} = {c=3}{},
  cell{1}{8} = {c=3}{},
  cell{2}{5} = {c=3}{},
  cell{2}{8} = {c=3}{},
  cell{9}{5} = {c=3}{},
  cell{9}{8} = {c=3}{},
  hlines,
  vline{2-9,9} = {1-2,9}{},
  vline{2-10} = {3-8}{},
}
{Formulation \& \\initial guess} & {Zero initial guess \\MPCC} & {Manual initial guess\\MPCC} & ADMM                & {Data-driven \\initial guess MPCC} &         &         & {Data-driven \\initial guess MIQP  } &        &         \\
Note                          & {Zero \\initial guess }     & {One Manual\\initial guess}  & {ADMM for \\MINLPs} & {KNN selects \\top 3 guesses}      &         &         & {CoCo select \\top 10 guesses  }    &        &         \\
Amount of data                & 0                           & 0                            & 0                   & 100                                & 500     & 1000    & 15000                               & 30000  & 80000   \\
Success Rate                  & <1\%                      & 78.25\%                      & 96.5\%              & 93.25\%                            & 97.75\% & 99.40\% & 94.50\%                              & 95.50\% & 98.75\% \\
Avg. Solve Time               & /                        & 29ms                         & 260ms               & 61ms                               & 30ms    & 28ms    & 65ms                                & 45ms   & 16ms    \\
Max. Solve Time               & /                        & 198ms                        & 1.29sec             & 257ms                              & 128ms   & 110ms   & 190ms                               & 182ms  & 174ms   \\
Avg. Objective                & /                       & -8620                        & -7250               & -7982                              & -8492   & -8506   & -8637                               & -8631  & -8607   \\
Avg. \# trials            & /                           & 1                            & 4.73                & 1.27                               & 1.05    & 1.02    & 4.85                                & 3.85   & 2.57    \\
Solver                        & Knitro                      & Knitro                       & Gurobi+Knitro       & Knitro                             &         &         & OSQP                                &        &         
\end{tblr}
\end{table*}

\subsection{Baselines}
\label{Sec:Baseline}
\begin{enumerate}
    \item \textit{Zero Initial Guesses} As a baseline, we solve the MPCC re-formulation with the default initial guess from the solvers (all zeros). The results are shown in the column named ``Zero initial guess MPCC'' of Table \ref{Tab:benchmark}. 
    \item \textit{Manually Designed Initial Guesses} Another baseline is to solve the MPCC re-formulation with manually designed initial guesses. The book placement requires inserting one book minimizing the movements of the existing books. Therefore, a good initial guess is to directly use the continuous and discrete variables from the original scene. We solve the separating planes using the original scene as initial guesses for $\boldsymbol{a}$ and $b$. The results are shown in column named ``Manual initial guess MPCC'' of Table \ref{Tab:benchmark}.
    \item \textit{ADMM for MINLPs} For the last non-data-driven baseline, we implemented ADMM method designed for MINLP described in Section Section \ref{Sec:ADMM_algorithm}. Manually designed initial guesses described above are used for the nonlinear sub-problem to increase the feasibility rate. Efforts are devoted to tuning the weights $\boldsymbol{G}$, $\gamma$, and scaling the variables properly to ensure the performance is on the better end. The results are shown in the column named ``ADMM'' of Table \ref{Tab:benchmark}.
\end{enumerate}

Despite the $\epsilon$ relaxation of complementary constraints, the rate of solving the problem successfully with zero warm-start is almost 0, revealing the challenge of solving MPCCs without good initial guesses. The performance is improved significantly with one manually designed initial guess, yet remains unsatisfactory. We also noticed that the success rates for manually designed initial guesses drop as the number of books increases. When there are four books on the shelf, inserting the 5th book has less than $60\%$ success rate. Therefore, this approach may not scale to problems with much larger sizes.

ADMM works fairly well for getting feasible solutions. However, it usually takes 4-5 iterations (1 iteration includes 1 MIQP and 1 NLP) to get a feasible solution, hence the solving speed is impaired. The nonlinear formulation takes more than 70\% of the solving time, due to the separating plane constraints. In addition, the average optimal cost from ADMM is harmed by the extra consensus terms in the objective function to guide the convergence.

\subsection{MPCC Re-formulation}
\label{Sec:Complementary_formulation}
We used the K-nearest neighbor algorithm to pick out the top three candidate solutions from the dataset of sizes 100, 500, and 1000, respectively. The results are shown in the column named ``Data-driven initial guess MPCC'' of Table \ref{Tab:benchmark}. A simple KNN algorithm with a few candidate solutions can significantly increase the rate of getting a feasible solution to more than $98\%$. The average trial of solving the problem is only slightly more than 1 showing that most problem instances are solved with only one trial. The average solving speed is also one order of magnitude faster than the non-data-driven ADMM algorithm.



\subsection{MIQP Re-formulation}
\label{Sec:MIP_formulation}


We trained the CoCo network to pick out the first 10 candidate binary solutions from datasets of sizes 15000, 30000, and 80000, respectively. The resultant QPs were then solved by OSQP, which used the previous solutions to warm-start the next problem for faster solving speeds. The results are shown in the column named ``Data-driven initial guess MIQP'' of Table \ref{Tab:benchmark}. Our results show that for this specific problem, the MIQP re-formulation achieves the fastest solving speed among the formulations tested. On the other hand, CoCo takes on more trials to find a feasible solution than the MPCC re-formulation. We also tested the KNN approach to select the first 10 candidate binary solutions and got similar results as using the CoCo network. 


\subsection{Analysis of Results}
\label{Sec:Comparison}
According to our test, data-driven methods can simultaneously achieve better optimality, faster solving speed, and higher success rates than non-data-driven methods for the book placement problem. The MPCC re-formulation and MIQP re-formulation differs mostly at the amount of data used to achieve similar solving speeds and success rates. The solving speeds heavily rely on the solver used. Since we do not have access to the details of numerical schemes inside KNITRO or Gurobi, we avoid reasoning in this aspect.

Our MIQP re-formulation achieves $>95\%$ success rate using a dataset about size 30000. For datasets less than 10000, CoCo has success rates generally $<90\%$. On the other hand, MPCC re-formulation achieves $>90\%$ success rates with the number of data on the magnitude of hundreds. We noticed that KNITRO rarely changes the initial binary guesses for $x[j], j \in \mathcal{J}$. Therefore, this difference comes from the bilinear constraints ~\eqref{eqn:formulation-4} converted into multiple linear envelopes with additional binary variables $\boldsymbol{\delta}$. For MIQP re-formulation, the learner directly provide $\boldsymbol{\delta}$ that fixes the intervals $[x^i_{p_s}, x^{i+1}_{p_s}]$ and $[x^j_{q_s}, x^{j+1}_{q_s}]$ for $x[p_s]$ and $x[q_s]$ by adding linear constraints. The QP solver finds $x[p_s]$ and $x[q_s]$ inside the intervals. Since $x[p_s]$ and $x[q_s]$ within the predicted intervals, if the learner provides the wrong intervals, the QP is infeasible, and the learner would need to try again. Sufficiently large dataset is required to train the learner to predict the correct intervals. However, for MPCC re-formulation, KNN provides initial guesses for $x[p_s]$, $x[q_s]$, then the NLP solver uses numerical schemes to satisfy ~\eqref{eqn:formulation-4} and other constraints. Since NLP solvers are designed to search and satisfy nonlinear constraints, even restore from infeasible initial guesses, the learner does not have to guarantee the feasibility of initial guesses as seen by smaller numbers of trials.

Roughly speaking, if one has a limited amount of data, our results support keeping the bilinear constraints ~\eqref{eqn:formulation-4} and letting the learner provide initial guesses of $x[p_s]$, $x[q_s]$ to the NLP solvers, instead of decomposing ~\eqref{eqn:formulation-4} into multiple linear envelopes like ~\eqref{eqn:sos2_envelopes} does and letting the learner select the correct intervals for $x[p_s]$, $x[q_s]$ to reside in.

\section{Conclusion, Discussion and Future Work} 
\label{Sec:conclusion}
This paper compares data-driven performances of MPCC or MIP re-formulations applied to the book placement problem formulated as a MIBLP.

Data-driven methods can simultaneously achieve better optimality, faster solving speed, and higher success rates than non-data-driven methods. The MPCC re-formulation has a higher chance of finding feasible solutions than the MIQP re-formulation from less amount of data. The MIQP re-formulation requires significantly more data to achieve similar performances as the MPCC re-formulation. 

An obvious challenge with the data-driven approach is how to generalize the training to cases outside the dataset. To achieve this goal, simple learners such as KNN do not perform well, and learning methods such as auto-encoders are to be explored.


{
\bibliographystyle{IEEEtran}
\bibliography{references}
}

\end{document}